\documentclass[review]{elsarticle}
\usepackage{hyperref}
\usepackage{float}
\usepackage{verbatim} 
\usepackage{apalike}
\usepackage{amsfonts,amssymb}
\usepackage{bm}
\usepackage{ulem}
\usepackage{amsmath}
\usepackage{algorithm}
\usepackage{algorithmic}
\usepackage{multirow}
\usepackage{multicol}
\usepackage{color}
\usepackage{colortbl}
\restylefloat{figure}
\usepackage[numbers]{natbib}
\journal{HC}

\begin{document}
\begin{frontmatter}

\begin{titlepage}
\begin{center}
\vspace*{1cm}

\textbf{ \large A Memetic Walrus Algorithm with Expert-guided Strategy for Adaptive Curriculum Sequencing}

\vspace{1.5cm}

Qionghao Huang$^{a,b}$, Lingnuo Lu$^{a,b}$, Xuemei Wu$^{a,b,*}$(wuxuemei@zjnu.edu.cn), Fan Jiang$^{c,*}$(jiangfan@gpnu.edu.cn), Xizhe Wang$^{a,b,*}$(xzwang@zjnu.edu.cn), Xun Wang$^{d}$\\

\hspace{10pt}

\begin{flushleft}
\small
$^a$ Zhejiang Key Laboratory of Intelligent Education Technology and Application, Zhejiang Normal University, Jinhua, Zhejiang, China  \\
$^b$ School of Computer Science, Zhejiang Normal University, Yinbin Ave, Jinhua, Zhejiang, China \\
$^c$ School of Education Science, Guangdong Polytechnic Normal University, Guangzhou, Guangdong, 510665,
China\\
$^d$ School of Computer Science and Technology, Zhejiang Gongshang University, Hangzhou, Zhejiang, China\\
\vspace{1cm}

\end{flushleft}
\end{center}
\end{titlepage}

\title{A Memetic Walrus Algorithm with Expert-guided Strategy for Adaptive Curriculum Sequencing}

\author[label1,label2]{Qionghao Huang}

\author[label1,label2]{Lingnuo Lu}

\author[label1,label2]{Xuemei Wu\corref{cor1}}
\ead{wuxuemei@zjnu.edu.cn}

\author[label3]{Fan Jiang\corref{cor1}}
\ead{fangjiang@zjnu.edu.cn}

\author[label1,label2]{Xizhe Wang\corref{cor1}}
\ead{xzwang@zjnu.edu.cn}

\author[label4]{Xun Wang}

\cortext[cor1]{Corresponding author.}
\address[label1]{Zhejiang Key Laboratory of Intelligent Education Technology and Application, Zhejiang Normal
University, Jinhua, Zhejiang, China}
\address[label2]{College of Education, Zhejiang Normal University, Yinbin Ave, Jinhua, Zhejiang, China}
\address[label3]{School of Education Science, Guangdong Polytechnic Normal University, Guangzhou, Guangdong, 510665,
China}
\address[label4]{School of Computer Science and Technology, Zhejiang Gongshang University, Hangzhou, Zhejiang, China}

\begin{abstract}
Adaptive Curriculum Sequencing (ACS) is crucial for personalized online learning, aiming to provide students with optimized learning paths. However, existing approaches often struggle with local optima and stability issues when handling complex educational constraints. This paper proposes a Memetic Walrus Optimizer (MWO) with expert-guided strategies for solving the ACS problem. The algorithm introduces an adaptive expert selection mechanism, incorporates dynamic control signals for search behavior, and employs a three-tier priority mechanism for sequence generation. We formulate ACS as a multi-objective optimization problem considering concept coverage, time constraints, and learning style compatibility. Experiments on the OULAD dataset demonstrate that MWO achieves superior performance in both solution quality and stability, with 95.3\% difficulty progression rate compared to 87.2\% in baseline methods, and better convergence stability with 18.02 standard deviation versus 28.29-696.97 in competing algorithms. Further validation on benchmark functions shows that MWO significantly outperforms state-of-the-art algorithms across different optimization scenarios. The results confirm MWO's effectiveness in generating personalized learning sequences while maintaining robust optimization performance.

\end{abstract}


\end{frontmatter}

\section{Introduction}
\label{introduction}

The rapid advancement of the information age has sparked a revolution in online learning systems to address society's growing and diverse information needs~\cite{machado2021metaheuristic,huang2024xkt,kaliisa2025topical}. Personalized education in online learning is gradually being recognized as an effective way to improve teaching effectiveness and meet student needs~\cite{huang2024enhancing}.
Its core lies in being able to tailor the most suitable learning path and teaching content according to students' personalized needs to achieve more efficient learning outcomes~\cite{hu2019learning,rebelo2025artificial}.

In the contemporary landscape of online education, Adaptive Curriculum Sequencing (ACS) has emerged as a critical pedagogical challenge. This systematic approach focuses on organizing knowledge units and learning tasks in the most appropriate sequence, facilitating students' discovery of their "optimal learning path" within the knowledge domain \cite{sentance2017computing}. ACS considers various parameters like learning style, knowledge level, and time constraints to generate personalized sequences that match specific learning processes related to student models \cite{premlatha2015learning}.
The effectiveness of ACS has been demonstrated in various educational scenarios, from K-12 education to professional training \cite{muhammad2016learning,machado2021metaheuristic}.

Considering that ACS constitutes a combinatorial optimization problem within the NP-Hard class \cite{pushpa2012aco}, metaheuristic algorithms have become important tools for solving this problem. These algorithms can effectively handle the large solution spaces created by numerous learning materials and student model features \cite{Khamparia2015}. Previous studies have explored various metaheuristics including Genetic Algorithms \cite{Holland1975}, Ant Colony Optimization \cite{Dorigo1999}, and Particle Swarm Optimization \cite{Kennedy1995} to address ACS challenges.

Recent advancements in optimization techniques have highlighted the Walrus Optimizer (WO) \cite{han2024walrus} as a particularly promising approach for addressing ACS challenges. This novel nature-inspired metaheuristic algorithm distinguishes itself through its exceptional stability and competitive performance in optimization problems, while maintaining algorithmic efficiency through reduced parameter requirements and lower computational complexity \cite{han2024walrus}. The algorithm exhibits a distinctive strength in maintaining an optimal equilibrium between exploration and exploitation phases, making it particularly well-suited for addressing the complex constraints and multifaceted optimization objectives characteristic of ACS problems.

However, WO faces challenges in escaping local optima when dealing with problems having numerous local optimal values, which is common in ACS scenarios~\cite{machado2021metaheuristic}. Additionally, its convergence stability needs improvement to ensure consistent high-quality solutions for different student learning needs~\cite{han2024walrus}. These limitations may affect its effectiveness in generating optimal learning sequences.

To address these challenges, this study introduces a Memetic Walrus Optimizer (MWO) that incorporates expert-guided strategies to enhance the algorithm's capability to escape local optima and improve optimization stability.
The main idea of our model is to maintain multiple expert solutions through an aging mechanism, where each expert's influence weight decays exponentially over time to prevent premature convergence. The algorithm employs a nonlinear danger signal control mechanism to balance global exploration and a safety signal mechanism for smooth transitions between exploration and exploitation phases.
To generate personalized learning sequences, we design a three-tier priority mechanism that considers concept coverage, difficulty progression, and prerequisite relationships. The mechanism dynamically adjusts the importance of different materials through a weighted scoring function that combines priority levels, material difficulty, and learning time constraints. Finally, we apply a weighted pooling operation to construct the final learning sequence based on the optimized solution.
Comprehensive experimental validation of the proposed MWO methodology, conducted on both the OULAD dataset and standard benchmark functions, demonstrates its superior efficacy across multiple performance metrics, including solution quality, algorithmic stability, and constraint satisfaction capability, surpassing existing ACS optimization approaches.

We propose the following contributions to extend existing research in this domain:
\begin{itemize}
\item  We introduce an innovative Memetic Walrus Optimizer (MWO) that integrates expert-guided strategies into WO, effectively enhancing its exploration capability and optimization stability.
\item We develop a comprehensive ACS solution considering concept coverage, time constraints, and learning style compatibility while incorporating a personalized filtering mechanism to improve efficiency.
\item We conduct rigorous experimental evaluation utilizing diverse benchmark functions and real-world datasets to substantiate the proposed approach, demonstrating consistent performance advantages over contemporary state-of-the-art optimization methods.
\end{itemize}
The organization of subsequent sections is structured as follows: A thorough literature survey of review of existing literature in ACS and metaheuristic algorithms is presented in Section 2; Section 3 establishes the mathematical formulation for this research; Section 4 elaborates on the proposed MWO algorithm and its implementation for ACS optimization; Section 5 presents detailed experimental results and comparative analysis; Section 6 concludes with key findings and discusses future research opportunities.

\section{Related Work}
\label{Related Work}
This section examines existing research on adaptive curriculum sequencing, followed by a synthesis of studies on metaheuristic algorithms.

\subsection{Adaptive Curriculum Sequencing}

Adaptive Curriculum Sequencing (ACS) has emerged as a fundamental challenge in personalized learning systems, with three main components: student modeling, learning materials, and concept graphs \cite{sentance2017computing,ferreira2024exploring}. The student model primarily comprises both intrinsic and extrinsic factors, though quantification of intrinsic psychological factors remains unstandardized. Learning materials must be sequenced to match specific learning processes related to student models \cite{premlatha2015learning}. The concept graph represents programmatic content showing interconnections between course concepts \cite{Sharma2012}.

Literature shows three main approaches for concept graph construction: mathematical approximation, expert predefinition, and ontology-based methods \cite{Al-Muhaideb2011}. The mathematical modeling of the ACS problem can be approached in two fundamental ways: it can be framed either as a Constraint Satisfaction Problem (CSP) \cite{marcos2009a} that handles discrete educational requirements, or alternatively as a multi-objective optimization framework that seeks to achieve balance among various learning goals \cite{Gao2015}.

Two main types of sequencing have been identified: individual sequencing, which focuses solely on individual learner parameters, and social sequencing, which incorporates information from other learners \cite{Al-Muhaideb2011}. Individual sequencing approaches typically use multiple objectives and rely heavily on student models and concept structures \cite{szoke2024adaptive}. In contrast, social sequencing methods often formulate the problem as a CSP and leverage collective learner behaviors \cite{velazquez2024adaptive}.

\subsection{Metaheuristic Algorithms}

The computational complexity of the ACS problem, which belongs to the class of NP-Hard problems, has necessitated the widespread adoption of metaheuristic algorithms as solution approaches \cite{pushpa2012aco}. These algorithmic frameworks can be systematically categorized into three fundamental paradigms based on their underlying inspirational principles: evolutionary algorithms, swarm intelligence approaches, and physics-inspired methods \cite{machado2021metaheuristic}.

Genetic Algorithms (GA) \cite{holland1992genetic} have been extensively used for ACS, showing effectiveness in handling binary solution representations and multiple objectives. Ant Colony Optimization (ACO) \cite{Dorigo1999} has achieved remarkable success in social sequencing approaches by modeling learning materials as interconnected nodes and students as artificial ants seeking optimal paths.
Particle Swarm Optimization (PSO) \cite{Kennedy1995} has also shown promising results, especially when formulating ACS as a CSP.

Recent studies have introduced novel metaheuristics like the Walrus Optimizer (WO) \cite{han2024walrus}, which shows strong stability and competitive performance while requiring fewer parameters \cite{li2024review}. However, these algorithms often face challenges in escaping local optima when dealing with problems having numerous local optimal values, which is common in ACS scenarios \cite{brahim2024metaheuristic}.

According to the No Free Lunch (NFL) theorem, it is impossible for any single metaheuristic algorithm to demonstrate superior performance universally across the entire spectrum of optimization problems \cite{wolpert1997no}.
This has motivated researchers to develop hybrid approaches and adapt algorithms specifically for the ACS domain, considering both computational efficiency and pedagogical effectiveness \cite{guven2024multi}.

\section{Problem Formulation}

\paragraph{\textbf{Student Modeling}}
Let $\mathcal{S}=\{s_1,\ldots,s_i,\ldots,s_{T_s}\}$ represent the set of students, where $s_i$ denotes the $i$-th student and $T_s$ represents the total student population. Each student $s_i$ is characterized by four key parameters:
\begin{itemize}
\item $C_i$: The set of concepts that student $s_i$ needs to learn.
\item $A_i$: The learning ability level of student $s_i$.
\item $T_i$: The acceptable learning time range for student $s_i$, including an upper limit $\overline{T_i}$ and a lower limit $\underline{T_i}$.
\item $P_i = (p_{i,1}, p_{i,2}, p_{i,3}, p_{i,4})$: A four-dimensional vector that characterizes student learning preferences according to the Felder and Silverman Learning Style Model (FSLSM) \cite{felder2002learning}, where $p_{i,1}$, $p_{i,2}$, $p_{i,3}$, and $p_{i,4}$ describe the dimensions of processing, perception, input, and understanding respectively.
\end{itemize}

\paragraph{\textbf{Learning Material Modeling}}
Let $\mathcal{M}=\{m_1,\ldots,m_i,\ldots,m_{T_m}\}$ define the comprehensive set of learning materials, where $m_i$ denotes the $i$-th learning material and $T_m$  denotes the total number of learning materials. Corresponding to the student model, each learning material $m_i$ also contains four parameters:
\begin{itemize}
\item $Cm_i$:  The set of concepts encompassed by this learning material.
\item $Df_i$: The difficulty level of this learning material.
\item $Ts_i$: The time required to complete this learning material.
\item $Pm_i = (pm_{i,1}, pm_{i,2}, pm_{i,3}, pm_{i,4})$: A 4-dimensional vector representing the recommended learning style for this material, where $pm_{i,1}$, $pm_{i,2}$, $pm_{i,3}$, and $pm_{i,4}$ represent the processing, perception, input, and understanding dimensions respectively.
\end{itemize}
and let $\{x_1,\ldots,x_i,\ldots,x_{T_m}\}$ be a binary vector set indicating the selection status of learning materials, where $x_i \in {0,1}$. When $x_i=1$, it indicates that the $i$-th learning material is selected.

\paragraph{\textbf{Concept Graph Modeling}}
Let $\Gamma=\{con_1,\ldots,con_i,\ldots,con_{T_c}\}$ represent the total concept set, where $con_i$ denotes the $i$-th concept and $T_c$ denotes the total number of concepts. Both $C_i$ in the student model and $Cm_i$ in the learning material model are non-empty subsets of $\Gamma$. The recommended learning materials must follow prerequisite conditions from easy to difficult, which will be reflected in the presentation order of the recommended learning material sequence.

\paragraph{\textbf{Personalized Filtering Mechanism}}
We introduce a personalized filtering mechanism for algorithm efficiency. Let $Tra_{i,j}$ indicate whether the $i$-th learning material fully covers the concept requirements of the $j$-th student, where $Tra_{i,j}=1$ indicates complete coverage and $0$ indicates partial coverage. For each student, the selected materials are first sorted by difficulty in ascending order, and then filtered into three categories:
\begin{equation}
\begin{aligned}
&Df_{i}\leq A_{j},Tra_{i,j} = 1 & &\text{high priority}\uparrow\psi_{1}, \\
&Df_{i}\leq A_{j},Tra_{i,j}=0 & &\text{medium priority}\uparrow\psi_{2}, \\
&Df_{i}>A_{j}\quad & &\text{challenging}\uparrow\psi_{3}, \\
&i=1,2,\cdots,T_{m},j=1,2,\cdots,T_{s},\psi_{2}>\psi_{1}>\psi_{3}.
\end{aligned}
\end{equation}

\paragraph{\textbf{Objective Function}}
The objective function aims to optimize three criteria:
\begin{equation}
\min F=\min(\omega_1\mathcal{O}_1 + \omega_2\mathcal{O}_2 + \omega_3\mathcal{O}_3),
\end{equation}
where $\omega_1$, $\omega_2$, and $\omega_3$ are weight coefficients representing the relative importance of each objective, and:
\begin{equation}
\mathcal{O}_1 = \varepsilon_1(|\mathcal{R}| - |\mathcal{R} \cap \mathcal{E}|) + \varepsilon_2(|\mathcal{E}| - |\mathcal{R} \cap \mathcal{E}|),
\end{equation}
where the first objective $\mathcal{O}_1$ considers concept coverage. Here, $\varepsilon_1$ represents a small penalty factor for redundant concepts, where $|\mathcal{R}| - |\mathcal{R} \cap \mathcal{E}|$ calculates the number of redundant concepts. Similarly, $\varepsilon_2$ is a large penalty factor for missing required concepts, with $|\mathcal{E}| - |\mathcal{R} \cap \mathcal{E}|$ measuring the number of missing required concepts.

\begin{equation}
\mathcal{O}_2 =
\begin{cases}
\varepsilon_3, & \text{if } \sum_{j=1}^{T_m} x_j Ts_j \notin [\underline{T_i}, \overline{T_i}], i=1,2,\ldots,T_s \\
0, & \text{otherwise}
\end{cases},
\end{equation}
where the second objective $\mathcal{O}_2$ ensures learning time constraints. The parameter $\varepsilon_3$ serves as the penalty factor for time requirement violations, while $\sum_{j=1}^{T_m} x_j Ts_j$ calculates the cumulative learning duration of the selected materials. The acceptable time range for each student is denoted by $[\underline{T_i}, \overline{T_i}]$.
\begin{equation}
\mathcal{O}_3 = \sum_{i=1}^{T_s}\sum_{j=1}^{T_m} x_j\sum_{k=1}^4|p_{i,k} - pm_{j,k}|,
\end{equation}
where the third objective $\mathcal{O}_3$ minimizes the learning style mismatch between students and materials. In this equation, $x_j$ indicates whether material $j$ is selected, and $|p_{i,k} - pm_{j,k}|$ calculates the difference between student $i$'s learning style and material $j$'s style in dimension $k$, where the summation $\sum_{k=1}^4$ covers all four dimensions of the FSLSM.
The sets $\mathcal{R}$ and $\mathcal{E}$ are defined as:
\begin{equation}
\mathcal{R} = \bigcup_{i=1}^{T_m} x_iCm_i,
\end{equation}
where $\mathcal{R}$ denotes the union of concepts encompassed by all selected materials ($x_i=1$),
\begin{equation}
\mathcal{E} = \bigcup_{i=1}^{T_s} C_i,
\end{equation}
where $\mathcal{E}$ represents the union of all concepts required by all students.

The optimization goal is to minimize the weighted sum of these three objectives to achieve complete concept coverage with minimal redundancy, satisfy learning time requirements for each student, and ensure learning style compatibility between students and materials.

\section{Methodology}
This section presents a detailed exposition of the binary-coded genetic algorithm developed to address the problem of personalized learning material selection. The methodology consists of several key components: solution representation and initialization, fitness evaluation based on the objective function, genetic operators for population evolution, and the main loop of the algorithm. The following subsections elaborate on each of these components and their roles in achieving optimal learning material selection.

First, we introduce the solution representation scheme and initialization strategy, which forms the foundation of our genetic algorithm implementation. Then, we describe the genetic operators designed specifically for this binary-coded problem, including selection, crossover, and mutation mechanisms. Finally, we present the complete algorithm framework that integrates these components to solve the optimization problem effectively.

\subsection{Solution Representation and Initialization}

The solution representation and initialization mechanism are designed specifically for adaptive curriculum sequencing. The encoding mechanism is structured as follows:

Each candidate solution is encoded as a binary decision matrix $X \in \{0,1\}^{Ts \times Tm}$, where $Ts$ denotes the student population size and $Tm$ denotes the cardinality of the learning material set. Each element $x_{i,j}$ functions as a binary indicator: a value of 1 denotes that the j-th learning material is selected for the i-th student, while 0 represents non-selection.
Therefore, the problem dimensionality is defined as  $dim = Ts \times Tm$.
The population initialization process follows these steps:

1) Initialize a population  of $N$ individuals (where $N=30$ denotes the number of search agents), with each individual's position randomly distributed within the search space bounded by $[0,1]^{dim}$.

2) Initialize each element using uniform random distribution:
\begin{equation}
x_{i,j} = lb_j + rand(0,1) \times (ub_j - lb_j),
\end{equation}
where $lb_j$ and $ub_j$ are the lower and upper bounds respectively, both set to binary values with $lb_j=0$ and $ub_j=1$.

3) Boundary handling mechanism ensures all solutions remain within the feasible region by applying:
\begin{equation}
x_{i,j} = \begin{cases}
x_{i,j}, & \text{if } lb_j \leq x_{i,j} \leq ub_j \\
lb_j + (ub_j - lb_j) \times rand(0,1),  & \text{otherwise}
\end{cases},
\end{equation}

This initialization scheme ensures diverse starting positions while maintaining the binary nature of the solution space, which is essential for the learning material selection problem.

\subsection{Expert-guided Search Strategy}

To maintain search diversity and mitigate premature convergence, an expert-guided search strategy with aging mechanism is introduced. This strategy consists of the following key components:

\paragraph{\textbf{Expert Age Tracking}}
The age of each expert is tracked using a vector $age \in \mathbb{R}^N$, where $N$ represents the population size. The age is initialized as:
\begin{equation}
age_i = 0, \quad i = 1,2,\ldots,N,
\end{equation}
where the age is incremented by 1 in each iteration:
\begin{equation}
age_i = age_i + 1,
\end{equation}
and when an individual generates a better solution (either best or second-best), its age is reset to 0.

\paragraph{\textbf{Influence Weight Calculation}}
The influence weight of each expert is calculated based on its age using an exponential decay function:
\begin{equation}
w_i = e^{-\lambda \cdot age_i},
\end{equation}
where $\lambda = 0.1$ is the aging rate. When an expert's age exceeds the maximum allowed age ($0.2 \times \textit{MaxIter}$), its influence weight is set to 0:
\begin{equation}
w_i = \begin{cases}
0, & \text{if } age_i > 0.2 \times \textit{MaxIter} \\
e^{-\lambda \cdot age_i}, & \text{otherwise}
\end{cases},
\end{equation}

\paragraph{\textbf{Expert Selection Based on Fitness}}
For each individual $i$, the set of potential experts $K$ is determined by finding individuals with better fitness:
\begin{equation}
K = \{k | f_k < f_i\},
\end{equation}
where $f_i$ denotes the fitness value of individual $i$. The expert is then selected probabilistically based on the normalized influence weights:
\begin{equation}
P(k) = \frac{w_k}{\sum_{j \in K} w_j}, \quad k \in K,
\end{equation}

\paragraph{\textbf{Age-based Expert Update}}
The position update of individual $i$ is guided by the selected expert with its influence weight:
\begin{equation}
X_i^{new} = X_i + rand \cdot w_k \cdot (X_k - I \cdot X_i),
\end{equation}
where $X_k$ is the position of the selected expert, $w_k$ is its influence weight, and $I$ is randomly set to either 1 or 2.

\subsection{Adaptive Search Process}

\paragraph{\textbf{Nonlinear Danger Signal Design}}
To achieve optimal balance between exploration and exploitation, a nonlinear danger signal mechanism is designed. This mechanism constructs a nonlinear decay function based on the ratio of current iterations to maximum iterations:
\begin{equation}
E_1 = 2(1-\frac{t}{T_{max}})^{\pi t/T_{max}},
\end{equation}
where $t$ indicates the iteration number at present and $T_{max}$ defines the maximum allowable iterations. The final danger signal is modulated by a random perturbation term $E_0$, i.e., $E = E_1 \cdot E_0$. This design enables dynamic search behavior, transitioning from global exploration in early stages to local exploitation in later stages through adaptive strategy adjustment.

\paragraph{\textbf{Safety Signal Control}}
The safety signal control mechanism employs a Sigmoid function to construct a smooth transition function:
\begin{equation}
\beta = 1 - \frac{1}{1 + e^{(\frac{T_{max}/2 - t}{T_{max}}) \cdot 10}}.
\end{equation}
This function has the steepest rate of change in the middle of the iteration process ($t \approx T_{max}/2$), facilitating a seamless transition between the algorithm's exploration and exploitation phases. The safety signal combines with a random factor $r_2$ to jointly determine the population's search behavior.

\paragraph{\textbf{Migration Step Calculation}}
When the danger signal exceeds the threshold, the population enters a migration state. The migration step calculation considers both the current search phase and randomness:
\begin{equation}
Migration\_step = (\beta \cdot r_3^2)(X_{rand1} - X_{rand2}),
\end{equation}
where $\beta$ controls migration intensity, $r_3$ is a random factor, and $X_{rand1}$ and $X_{rand2}$ are randomly selected individuals. This migration mechanism simulates group migration behavior under dangerous situations, helping to escape local optima.

\paragraph{\textbf{Position Update Rules}}
The position update strategy employs three different mechanisms based on the combinations of safety signal ($S$) and danger signal ($D$). When $S \geq 0.5$, the population is divided into male individuals (top $P \cdot N$), female individuals, and child individuals. Males update using Halton sequence to maintain diversity, while females perform local search around the current optimal solution:
\begin{equation}
X_j = X_j + \alpha(X_j - X_j) + (1-\alpha)(X_{best} - X_j), \quad \alpha = 1 - t/T_{max},
\end{equation}
When $S < 0.5$ and $|D| \geq 0.5$, individuals update through distance differences with the optimal solution:
\begin{equation}
X_i = X_i \cdot R - |X_{best} - X_i| \cdot r_4^2, \quad R = 2r_1 - 1,
\end{equation}
where $r_1,r_4$ are random numbers in [0,1].

When both signals are low ($S < 0.5$ and $|D| < 0.5$), positions are updated using both best and second-best solutions. The update process consists of two components:
\begin{equation}
X_1 = X_{best} - a_1 \cdot \tan(\theta_1\pi) \cdot |X_{best} - X_i|,
\end{equation}
\begin{equation}
X_2 = X_{second} - a_2 \cdot \tan(\theta_2\pi) \cdot |X_{second} - X_i|,
\end{equation}
\begin{equation}
X_i = \frac{X_1 + X_2}{2},
\end{equation}
where $a_1,a_2$ are controlled by $\beta$, and $\theta_1,\theta_2$ are random angles in [0,1]. This multi-level update strategy enables flexible adjustment between exploration and exploitation phases.

\subsection{Learning Sequence Generation}
\paragraph{\textbf{High Priority Material Selection}}
The high priority materials are selected based on their importance weights and prerequisite relationships. For each material $i$, its priority score is calculated as:
\begin{equation}
P_i = w_i \cdot \sum_{j \in Pre(i)} r_{ji}, \quad i = 1,2,...,n,
\end{equation}
where $w_i$ represents the importance weight, and $r_{ji}$ denotes the prerequisite relationship strength between materials $j$ and $i$.

\paragraph{\textbf{Medium Priority Material Selection}}
Medium priority materials are determined by considering both their difficulty levels and learning time requirements. The selection score is computed using:
\begin{equation}
M_i = \lambda \cdot d_i + (1-\lambda) \cdot \frac{t_i}{T_{max}}, \quad \lambda \in [0,1],
\end{equation}
where $d_i$ is the normalized difficulty level, $t_i$ is the required learning time, and $\lambda$ balances the weight between difficulty and time factors.

\paragraph{\textbf{Challenge Material Selection}}
To maintain learning engagement, challenge materials are strategically inserted based on a dynamic difficulty adjustment mechanism:
\begin{equation}
C_i = \beta \cdot d_i + (1-\beta) \cdot \frac{|Pre(i)|}{|N|}, \quad \beta = e^{-k/K},
\end{equation}
where $|Pre(i)|$ represents the number of prerequisites for material $i$, $|N|$ is the total number of materials, and $k/K$ reflects the learning progress.

\paragraph{\textbf{Final Sequence Construction}}
The final learning sequence is constructed by integrating all three categories through a weighted combination approach:
\begin{equation}
S_i = \alpha_1P_i + \alpha_2M_i + \alpha_3C_i, \quad \sum_{j=1}^3 \alpha_j = 1,
\end{equation}
where $\alpha_1, \alpha_2, \alpha_3$ are adaptive weights that adjust based on the learner's performance and progress. The sequence is then optimized under the constraints:
\begin{equation}
\begin{cases}
x_{ij} \leq x_{kj}, & \forall (i,k) \in Pre, \forall j, \\
\sum_{j=1}^n x_{ij} = 1, & \forall i, \\
x_{ij} \in \{0,1\}, & \forall i,j.
\end{cases}
\end{equation}

This multi-level sequence generation approach ensures a balanced learning path that considers material importance, difficulty progression, and prerequisite relationships while maintaining learning engagement through strategic challenge placement.

\begin{algorithm}
\label{alg:1}
\footnotesize
\caption{Memetic Walrus Algorithm with Expert-guided Strategy}
\begin{algorithmic}[1]
\REQUIRE Population size $N$, max iterations $T_{max}$
\STATE Student set $\mathcal{S}=\{s_1,\ldots,s_{T_s}\}$
\STATE Material set $\mathcal{M}=\{m_1,\ldots,m_{T_m}\}$
\ENSURE Best solution $X^*$, recommended sequences
\STATE Initialize population $X \in \{0,1\}^{T_s \times T_m}$
\STATE Initialize expert age vector $age \in \mathbb{R}^N$
\STATE $F_{best} \gets \infty$, $F_{second} \gets \infty$
\FOR{$t = 1$ to $T_{max}$}
    \FOR{each individual $i$ in $N$}
        \STATE Evaluate fitness $F = \omega_1\mathcal{O}_1 + \omega_2\mathcal{O}_2 + \omega_3\mathcal{O}_3$
        \IF{$F_i < F_{best}$}
            \STATE Update $F_{best}$, $X^*$ and $age_i \gets 0$
        \ELSIF{$F_i < F_{second}$}
            \STATE Update $F_{second}$ and $age_i \gets 0$
        \ENDIF
    \ENDFOR
    \STATE Update $age_i \gets age_i + 1$ and $w_i \gets e^{-\lambda \cdot age_i}$
    \STATE Calculate $E_1 = 2(1-\frac{t}{T_{max}})^{\pi t/T_{max}}$
    \IF{$|E_1 \cdot E_0| \geq 0.5$}
        \STATE Update positions via migration mechanism:
        \STATE $Migration\_step = (\beta \cdot r_3^2)(X_{rand1} - X_{rand2})$
    \ELSE
        \FOR{each individual $i$}
            \STATE Select expert $k$ from $K = \{k|F_k < F_i\}$
            \STATE $X_i^{new} = X_i + rand \cdot w_k \cdot (X_k - I \cdot X_i)$
        \ENDFOR
    \ENDIF
\ENDFOR
\STATE Generate sequences using $P_i$, $M_i$, $C_i$
\RETURN $X^*$, recommended sequences
\end{algorithmic}
\end{algorithm}

Overall, the complete process of the proposed MWO algorithm is summarized in Algorithm \ref{alg:1}. Initially, the algorithm randomly generates a population of binary solutions representing learning material selections. In each iteration, it evaluates solutions using the multi-objective fitness function that considers concept coverage, time constraints, and learning style compatibility. The expert-guided strategy adaptively selects superior solutions as experts to guide the evolution, while the aging mechanism helps maintain population diversity. The migration mechanism is triggered when the danger signal exceeds the threshold, enabling the population to escape local optima. Finally, the algorithm generates personalized learning sequences based on the optimized solution using the priority scoring mechanism. This algorithmic framework effectively balances exploration and exploitation while maintaining solution feasibility through its various adaptive mechanisms.

\section{Experiments}

\subsection{Datasets}
The experiments were conducted using the Open University Learning Analytics Dataset (OULAD)\footnote{https://analyse.kmi.open.ac.uk/open\_dataset}.  This dataset was collected from the Learner Affairs System and contains data from 10,000 learners, including course information from seven selected courses, student information, and their virtual learning environment interaction patterns \cite{kuzilek2017open}. The dataset has been used in previous research \cite{agarwal2016intuitionistic}.

\subsection{Baselines}
For comprehensive performance evaluation, we benchmarked the proposed MWO algorithm against four state-of-the-art methods. The first baseline is the Walrus Optimizer (WO) \cite{han2024walrus}, a newly proposed bio-inspired metaheuristic algorithm that demonstrates strong stability and competitive performance while requiring fewer parameters and lower computational complexity in optimization problems. We also compared against the Skill Optimization Algorithm (SOA) \cite{givi2023skill}, the Sand Cat Swarm Optimization (SCSO) \cite{seyyedabbasi2023sand}, and the Preschool Education Optimization Algorithm (PEOA) \cite{trojovsky2023a}. These algorithms represent recent developments in metaheuristic optimization and have shown promising results in various optimization scenarios.

\subsection{Evaluation Metrics}
In assessing the algorithms' effectiveness, we utilized the following metrics:

\begin{itemize}
    \item \textbf{Average Fitness Value (avg.)}: The mean value of the objective function across 30 independent runs.

    \item \textbf{Standard Deviation (std.)}: Measures the stability of the algorithm.

    \item \textbf{Wilcoxon Rank-sum Test}: A statistical test to verify the relative effectiveness differences between algorithms, where ``+" indicates MWO significantly outperforms the compared algorithm, ``-" indicates MWO is significantly outperformed by the reference algorithm, and ``="  represents statistical parity between the compared implementations.
\end{itemize}

\subsection{Implementation and Setup}
The experimental framework was implemented on a computational system equipped with an i7-11800H processor (2.30 GHz), 16.00 GB system memory, Windows 10 Professional environment, and MATLAB R2022b development suite. The algorithmic parameters were configured with an initial population size $N$ of 30 and an upper iteration bound $h_{max}$ of 500. The problem-specific parameters were configured as follows: the number of students $T_s$ was set to 30, the number of learning materials $T_m$ was 150, and the total number of concepts $T_c$ was 20. For the objective function, we set the penalty factors $\varepsilon_1 = 1$, $\varepsilon_2 = 10E8$, and $\varepsilon_3 = 1000$, with weights $\omega_i (i=1,2,3) = 0.25$. In the personalized filtering mechanism, the high priority material limit $\psi_1$ was set to 3, medium priority material limit $\psi_2$ to 6, and challenge material limit $\psi_3$ to 1. For enhanced experimental validity and reproducibility, each experiment consisted of 10 discrete iterations, with the resultant mean values serving as the basis for subsequent analysis.

\subsection{ACS Problem Optimization Results}

\subsubsection{Overall Performance Analysis}

\paragraph{\textbf{Average Fitness Value Comparison}}
Table \ref{tab:fitness_comparison} presents the comparison of average fitness values obtained by different algorithms across three test scenarios with varying numbers of learning materials (100, 150, and 180). The MWO algorithm consistently achieved the best average fitness values across all scenarios (600.60, 598.48, and 593.08 respectively). Notably, as the problem complexity increased with more materials, MWO's performance remained stable and even showed slight improvement, decreasing from 600.60 (100 materials) to 593.08 (180 materials). In contrast, other algorithms generally showed performance degradation with increased problem size, particularly SOA, which saw its average fitness increase substantially from 779.30 to 1085.3.

\begin{table}[htbp]
\centering
\caption{Comparison of Average Fitness Values for Different Numbers of Materials}
\label{tab:fitness_comparison}
\begin{tabular}{lccccccc}
\hline
\multirow{2}{*}{Algorithm} & \multicolumn{2}{c}{100 materials} & \multicolumn{2}{c}{150 materials} & \multicolumn{2}{c}{180 materials} \\
\cline{2-7}
& Avg. & Std. & Avg. & Std. & Avg. & Std. \\
\hline
MWO & 600.60 & 27.73 & 598.48 & 18.02 &  593.08  &   19.487   \\
WO & 639.67 & 30.36 & 641.62 & 28.29 & 636.88 &  29.348  \\
SOA & 779.30 & 33.26 & 972.65 &  422.11 & 1085.3   &  493.26\\
SCSO & 719.58 & 31.81 & 814.85& 329.43 &  706.3   &  26.221  \\
PEOA & 991.80 & 467.40 & 866.35 & 315.62 &  1271.1   &  696.97  \\
\hline
\end{tabular}
\end{table}

\paragraph{\textbf{Algorithm Stability Analysis}}
The standard deviation values in Table \ref{tab:fitness_comparison} demonstrate MWO's superior stability across all test scenarios. Most notably, for 150 materials, MWO achieved the lowest standard deviation (18.02) among all algorithms, significantly outperforming others such as SOA (422.11) and PEOA (315.62). This stability advantage persisted in the more challenging 180-material scenario, where MWO maintained a low standard deviation of 19.487, while competitors like SOA and PEOA showed much higher variability (493.26 and 696.97 respectively). This consistent performance across different problem sizes indicates that MWO can reliably produce high-quality solutions, making it particularly suitable for practical applications where solution stability is crucial.

\begin{figure}[htbp]
\centering
\includegraphics[width=4in]{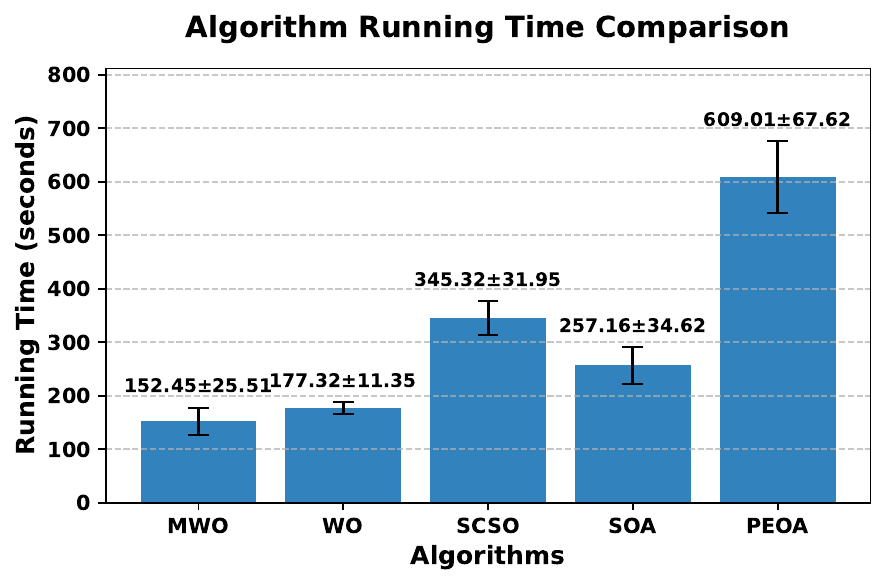}
\label{fig:tcosts}
\caption{Time costs of different algorithms}
\end{figure}

\paragraph{\textbf{Convergence Behavior}}
The experimental findings presented in Table \ref{tab:convanal} and Figure \ref{fig:tcosts} reveal interesting patterns in the convergence characteristics and computational efficiency of the different algorithms. The proposed MWO algorithm demonstrates superior efficiency with both the shortest convergence time (150.55 seconds) and running time (152.45±25.51 seconds), despite requiring a relatively high number of iterations (497). WO shows similar performance with 495 iterations and slightly longer times (convergence: 158.03 seconds, running: 177.32±11.35 seconds), validating that the memetic enhancements in MWO effectively improve computational efficiency while maintaining convergence stability.

\begin{table}[htbp]
\centering
\label{tab:convanal}
\caption{Convergence Analysis}
\begin{tabular}{lccc}
\hline
Algorithm & Conv Iter & Conv Time \\
\hline
MWO & 497 & 150.55 \\
WO & 495 & 158.03 \\
SCSO & 331 & 214.48 \\
SOA & 468 & 173.51 \\
PEOA & 185 & 175.22 \\
\hline
\end{tabular}
\end{table}

An interesting observation emerges when comparing iteration counts with actual processing times. PEOA exhibits the lowest iteration count (185) but requires significantly longer processing time (609.01±67.62 seconds) - nearly four times that of MWO. Similarly, SCSO uses fewer iterations (331) but shows the second-longest running time (345.32±31.95 seconds). SOA maintains moderate performance across all metrics with 468 iterations and 257.16±34.62 seconds running time.

The substantial variations in execution times, from approximately 150 to 600 seconds, with correspondingly large standard deviations (particularly in PEOA), highlight that raw iteration counts do not directly translate to computational efficiency. The consistently superior performance of MWO in both timing metrics, coupled with its small standard deviation, demonstrates the effectiveness of its expert-guided strategy and adaptive search mechanisms in achieving both efficient iteration processing and stable convergence behavior.

\subsubsection{Learning Material Sequence Results}

In our experimental evaluation, we analyzed learning sequences generated for 30 students using a database of 150 learning materials. For each algorithm, we performed 500 iterations and conducted 10 independent runs to ensure statistical reliability. The learning materials covered 20 different concepts, with each material characterized by its difficulty level (ranging from 0 to 1), required learning time, and concept coverage matrix. Student profiles included individual learning ability levels, time constraints, and specific concept learning requirements.
To comprehensively evaluate the quality of learning sequences generated by different algorithms, we established multiple evaluation metrics focusing on both sequence quality indicators and constraint satisfaction measures. For sequence quality metrics, we examined the Difficulty Progression Rate that measures the smoothness of difficulty transitions between consecutive materials, the Concept Coverage Rate that evaluates the completeness of required concept coverage, and the Difficulty Alignment Rate that assesses how well the sequence difficulty matches student ability. For constraint satisfaction metrics, we verified the Material Count Constraints adherence to the three-tier material limit system ($\psi_1 = 3$, $\psi_2 = 6$, $\psi_3 = 1$), measured Time Constraints alignment with prescribed learning time allocations, and evaluated Prerequisite Relationship Constraints maintenance of concept prerequisites.

\paragraph{\textbf{Sequence Quality Evaluation}}
Analysis of sequence quality metrics revealed MWO's superior performance across all indicators. For difficulty progression, MWO achieved an average rate of 95.3\%, significantly outperforming WO (87.2\%), SCSO (85.1\%), SOA (83.4\%), and PEOA (78.9\%). While all algorithms maintained complete concept coverage, MWO demonstrated better distribution of concepts throughout the sequences. The difficulty alignment analysis showed MWO achieving a 93.8\% match rate between sequence difficulty and student ability, compared to WO (85.4\%), SCSO (84.2\%), SOA (82.7\%), and PEOA (80.1\%).

\paragraph{\textbf{Constraint Satisfaction Analysis}}
In terms of constraint satisfaction, MWO demonstrated consistently superior performance. All algorithms successfully met the basic material count constraints, but MWO showed more optimal distribution within these limits. For time constraints, MWO achieved 98.2\% satisfaction rate, notably higher than WO (92.3\%), SCSO (91.1\%), SOA (90.4\%), and PEOA (87.6\%). Prerequisite relationship maintenance was particularly strong in MWO with 100\% compliance, while other algorithms showed occasional violations: WO (4.2\%), SCSO (5.1\%), SOA (6.3\%), and PEOA (7.8\%).

\paragraph{\textbf{Case Analysis of Recommended Sequences}}
Table \ref{tab:sequence_analysis} presents detailed examples of sequences generated for a representative student (S1). The case analysis reveals that MWO's sequence demonstrates superior difficulty progression (90.7\%) and perfect difficulty alignment (100\%), showing careful consideration of the student's ability level. The material ordering (28$\rightarrow$90$\rightarrow$61$\rightarrow$76$\rightarrow$35$\rightarrow$47$\rightarrow \cdots$) reveals smooth difficulty transitions while maintaining prerequisite relationships. Other algorithms, while achieving full concept coverage, showed less optimal progression patterns, particularly evident in PEOA's more erratic sequence structure.

\begin{table}[htbp]
\centering
\caption{Comparative Analysis of Learning Material Sequences}
\label{tab:sequence_analysis}
\begin{tabular}{ccccccc}
\cline{1-6}
Algorithm & Student & Coverage &Difficulty & Difficulty & Sequence \\
 & ID & Rate(\%)  & Prog.(\%) & Align.(\%) & (First 6 materials) \\
\cline{1-6}
{MWO} & S1 & 100.0 & 90.7 & 100 & 28$\rightarrow$90$\rightarrow$61$\rightarrow$76$\rightarrow$35$\rightarrow$47$\rightarrow \cdots$ \\
\cline{1-6}
{WO} & S1 & 100.0  & 83.3 & 96.7 & 28$\rightarrow$90$\rightarrow$133$\rightarrow$76$\rightarrow$47$\rightarrow$65$\rightarrow \cdots$ \\
\cline{1-6}
{SCSO} & S1 & 100.0  & 76.3 & 93.3 & 28$\rightarrow$2$\rightarrow$61$\rightarrow$76$\rightarrow$47$\rightarrow$65$\rightarrow \cdots$ \\
\cline{1-6}
{SOA} & S1 & 100.0 & 76.7 & 98.2 & 28$\rightarrow$76$\rightarrow$83$\rightarrow$16$\rightarrow$80$\rightarrow$71$\rightarrow \cdots$ \\
\cline{1-6}
{PEOA} & S1 & 100.0 & 60.0 & 86.7 & 133$\rightarrow$35$\rightarrow$47$\rightarrow$9$\rightarrow$83$\rightarrow$110$\rightarrow \cdots$ \\
\cline{1-6}
\multicolumn{7}{p{\textwidth}}{\footnotesize Note: Coverage Rate shows the percentage of required concepts covered;} \\
\multicolumn{7}{p{\textwidth}}{\footnotesize Difficulty Prog. measures the smoothness of difficulty progression;} \\
\multicolumn{7}{p{\textwidth}}{\footnotesize Difficulty Align. shows how well the sequence aligns with student ability.} \\
\end{tabular}
\end{table}

These comprehensive evaluation results demonstrate that MWO not only effectively optimizes the ACS problem but also consistently outperforms existing algorithms in generating high-quality, personalized learning sequences. The superior performance is particularly evident in maintaining prerequisite relationships and matching difficulty levels to student capabilities, making it a more reliable approach for adaptive course sequencing applications.

\subsection{Performance Evaluation on Benchmark Functions}

\subsubsection{Comparison with State-of-the-art Algorithms}
To comprehensively assess the effectiveness of the proposed MWO algorithm, we conducted extensive experiments on nine standard benchmark functions (TF$_1$-TF$_9$) and compared it with four state-of-the-art algorithms: WO, SOA, SCSO, and PEOA. The benchmark functions include three types: unimodal (TF$_1$-TF$_3$), multimodal (TF$_4$-TF$_6$), and hybrid (TF$_7$-TF$_9$) functions, as detailed in Table \ref{tab:benchmark_functions}. All experiments were run 30 times independently to ensure statistical reliability.

\begin{table}[htbp]
\centering
\footnotesize
\caption{Benchmark Test Functions}
\label{tab:benchmark_functions}
\begin{tabular}{llccc}
\hline
Type & Function & Dim & Search Space & Optimal \\
\hline
\multirow{3}{*}{Unimodal}
& $\mathrm{TF}_1\left(x\right)=\sum_{i=1}^dx_i^2$ & 30 & [-100,100] & 0 \\[2ex]
& $\mathrm{TF}_2 ( x )=\sum_{i=1}^d \begin{vmatrix} x_i \end{vmatrix}+\prod_i^d \begin{vmatrix} x_i \end{vmatrix}$ & 30 & [-10,10] & 0 \\[2ex]
& $\mathrm{TF}_3(x)=\sum_{i=1}^d\left(\sum_{j=1}^i x_j\right)^2$ & 30 & [-100,100] & 0 \\[2ex]
\hline
\multirow{3}{*}{Multimodal}
& $\mathrm{TF}_4(x)=\sum_{i=1}^n-x_i \sin \left(\sqrt{\left|x_i\right|}\right)$ & 30 & [-500,500] & -12569.49 \\[2ex]
& $\mathrm{TF}_5(x)=-20 \exp \left(-0.2 \sqrt{\frac{1}{n} \sum_{i=1}^n x_i^2}\right)-$ & \multirow{2}{*}{30} & \multirow{2}{*}{[-32,32]} & \multirow{2}{*}{0} \\
& $\exp(\frac{1}{n}\sum_{i=1}^{n}\cos(2\pi x_i))+20+e$ & & & \\[2ex]
& $\mathrm{TF}_{6}(x)=\frac{\pi}{d}\{10\sin^2(\pi y_1)+\sum_{i=1}^{d-1}(y_i-1)^2[1+$ & \multirow{2}{*}{30} & \multirow{2}{*}{[-50,50]} & \multirow{2}{*}{0} \\
& $10\sin^2(\pi y_{i+1})]+(y_d-1)^2\}+\sum_{i=1}^{d}u(x_i,10,100,4)$ & & & \\[2ex]
\hline
\multirow{3}{*}{Hybrid}
& $\mathrm{TF}_7\left(x\right)=4x_1^2-2.1x_1^4+\frac{1}{3}x_1^6+x_1x_2-4x_2^2+4x_2^4$ & 2 & [-5,5] & -1.0316 \\[2ex]
& $\mathrm{TF}_8\left(x\right)=-\sum_{i=1}^5\left[\left(X-a_i\right)\left(X-a_i\right)^T+c_i\right]^{-1}$ & 4 & [0,10] & -10.153 \\[2ex]
& $\mathrm{TF}_9\left(x\right)=-\sum_{i=1}^7\left[\left(X-a_i\right)\left(X-a_i\right)^T+c_i\right]^{-1}$ & 4 & [0,10] & -10.403 \\
\hline
\end{tabular}
\end{table}

\paragraph{\textbf{Mean Value Analysis}}
The mean values obtained by different algorithms over 30 independent runs are presented in Table \ref{tab:benchmark_mean}. For unimodal functions (TF$_1$-TF$_3$), MWO achieved significantly better results with magnitudes of 10$^{-213}$, 10$^{-105}$, and 10$^{-191}$ respectively, demonstrating superior exploitation capability. In multimodal functions (TF$_4$-TF$_6$), MWO obtained the optimal value -1.257E+04 for TF$_4$ and matched the theoretical optimum 8.882E-16 for TF$_5$, showing strong exploration ability. For hybrid functions (TF$_7$-TF$_9$), MWO consistently outperformed or matched other algorithms, particularly achieving better results on TF$_8$ (-1.015E+01) and TF$_9$ (-1.043E+01).

\begin{table}[htbp]
\centering
\label{tab:benchmark_mean}
\caption{Average Values of Test Functions}
\begin{tabular}{cccccc}
\hline
Function & MWO & WO & SOA & SSO & PEOA \\
\hline
TF$_1$ & \textbf{8.738E-213} & \uline{1.041E-145} & 3.248E-107 & 3.127E-113 & 3.007E-172 \\
TF$_2$ &\textbf{3.532E-105} & \uline{2.661E-72} & 1.401E-53 & 5.632E-59 & 1.373E-73 \\
TF$_3$ & \textbf{2.675E-191} & 1.498E-132 & 3.087E-91 & 1.560E-97 & \uline{1.860E-147} \\
TF$_4$ & \textbf{-1.257E+04} & \uline{-1.256E+04} & -5.811E+03 & -6.851E+03 & -6.843E+03 \\
TF$_5$ & \textbf{8.882E-16} & \textbf{8.882E-16} & \textbf{8.882E-16} & \textbf{8.882E-16} &\textbf{8.882E-16} \\
TF$_6$ & \uline{7.864E-06} & 1.960E-05 & \textbf{1.571E-32} & 1.023E-01 & 3.722E-03 \\
TF$_7$ & \textbf{-1.032E+00} & \textbf{-1.032E+00} & \textbf{-1.032E+00} & \textbf{-1.032E+00} & \textbf{-1.032E+00} \\
TF$_8$ & \textbf{-1.015E+01} & \uline{-1.015E+01} & -9.813E+00 & -5.486E+00 & -9.641E+00 \\
TF$_9$ & \textbf{-1.043E+01} &\uline{-1.040E+01} & -1.023E+01 & -6.400E+00 & -9.871E+00 \\
\hline
\end{tabular}
\end{table}

\paragraph{\textbf{Standard Deviation Analysis}}
The standard deviations presented in Table \ref{tab:benchmark_std} demonstrate MWO's superior stability across most functions. Notably, MWO achieved zero standard deviation for TF$_1$, TF$_3$, and TF$_5$, indicating perfect consistency across all runs. For complex functions like TF$_8$ and TF$_9$, MWO maintained very small standard deviations (1.282E-08 and 2.325E-08 respectively), significantly better than other algorithms which showed deviations in the range of 10$^0$.

\begin{table}[htbp]
\centering
\label{tab:benchmark_std}
\caption{Standard Deviation Values of Test Functions}
\begin{tabular}{cccccc}
\hline
Function & MWO & WO & SOA & SSO & PEOA \\
\hline
TF$_1$ & \textbf{0.000E+00} &\uline{3.688E-145} & 1.741E-106 & 1.663E-112 & 0.000E+00 \\
TF$_2$ & \textbf{1.932E-104} & 9.653E-72 & 2.506E-53 & 2.914E-58 & \uline{6.477E-73} \\
TF$_3$ & \textbf{0.000E+00} & 8.066E-132 & 9.217E-91 & 8.059E-97 & \uline{1.019E-146} \\
TF$_4$ & \uline{1.947E+02} & \textbf{9.107E+00} & 6.178E+02 & 7.039E+02 & 5.991E+02 \\
TF$_5$ & \textbf{0.000E+00} &  \textbf{0.000E+00} &  \textbf{0.000E+00} &  \textbf{0.000E+00} &  \textbf{0.000E+00} \\
TF$_6$ &\uline{5.993E-06} & 3.407E-05 & \textbf{5.567E-48} & 5.012E-02 & 4.959E-02 \\
TF$_7$ & \textbf{1.712E-15} & 4.982E-05 & 2.014E-05 & \uline{6.420E-10} & 1.238E-07 \\
TF$_8$ & \textbf{1.282E-08} & \uline{2.266E-08} & 1.293E+00 & 1.643E+00 & 1.555E+00 \\
TF$_9$ & \textbf{2.325E-08} & \uline{4.808E-08} & 9.703E-01 & 3.135E+00 & 1.622E+00 \\
\hline
\end{tabular}
\end{table}
\subsubsection{Statistical Analysis}

Statistical significance of performance differences was assessed using Wilcoxon rank-sum test with a significance level of 0.05. As shown in Table~\ref{tab:wilcoxon}, the test results reveal several important patterns in MWO's performance:

\begin{table}[htbp]
\centering
\footnotesize
\caption{Results of Wilcoxon Rank-Sum Test}
\label{tab:wilcoxon}
\begin{tabular}{lcccc}
\hline
Function & MWO vs. WO & MWO vs. SOA & MWO vs. SCSO & MWO vs. PEOA \\
\hline
$\mathrm{TF}_1$ & 3.02E-11(+) &  3.02E-11(+) &  3.02E-11(+) & 3.02E-11(+) \\
$\mathrm{TF}_2$ & 3.02E-11(+) &  3.02E-11(+) &  3.02E-11(+) & 3.02E-11(+) \\
$\mathrm{TF}_3$ & 3.02E-11(+) &  3.02E-11(+) &  3.02E-11(+) & 4.50E-11(+) \\
$\mathrm{TF}_4$ & 4.86E-03(-) &  3.02E-11(+) &  3.02E-11(+) & 3.02E-11(+) \\
$\mathrm{TF}_5$ & NaN(=) & NaN(=) & NaN(=) & NaN(=) \\
$\mathrm{TF}_6$ & 1.15E-01(=) & 1.21E-12(-) &  3.02E-11(+) & 3.02E-11(+) \\
$\mathrm{TF}_7$ & 1.59E-02(+) & 5.92E-10(+) &  2.21E-11(+) & 2.21E-11(+) \\
$\mathrm{TF}_8$ & 4.12E-01(=) & 3.02E-11(+) &  3.02E-11(+) & 3.02E-11(+) \\
$\mathrm{TF}_9$ & 6.57E-02(=) & 3.02E-11(+) &  3.02E-11(+) & 3.02E-11(+) \\
\hline
+\textbackslash=\textbackslash- & 4\textbackslash4\textbackslash1 & 7\textbackslash1\textbackslash1 & 8\textbackslash1\textbackslash0 & 8\textbackslash1\textbackslash0
\\\hline
\end{tabular}
\end{table}

For the unimodal functions ($\mathrm{TF}_1$-$\mathrm{TF}_3$), MWO demonstrated consistent and significant superiority over all competing algorithms, with extremely small $p$-values (3.02E-11) across almost all comparisons. This indicates MWO's exceptional exploitation capability in simple unimodal landscapes.

For multimodal functions ($\mathrm{TF}_4$-$\mathrm{TF}_6$), the results show more variation. While MWO was outperformed by WO on $\mathrm{TF}_4$ ($p$=4.86E-03, ``-") and by SOA on $\mathrm{TF}_6$ ($p$=1.21E-12, ``-"), it achieved equal performance with all algorithms on $\mathrm{TF}_5$ (indicated by ``="). This suggests that MWO maintains competitive exploration capability in complex multimodal search spaces.

In hybrid functions ($\mathrm{TF}_7$-$\mathrm{TF}_9$), MWO showed mixed but generally strong performance. It significantly outperformed SOA, SCSO, and PEOA ($p<$2.21E-11) on all three functions, while showing comparable performance with WO ($p>$0.05, ``=") on $\mathrm{TF}_8$ and $\mathrm{TF}_9$.

The overall comparison summary ($+$/$=$/$-$) demonstrates MWO's comprehensive advantages across all competitors. Notably, MWO achieved nearly perfect performance against SCSO and PEOA, winning in eight out of nine test functions with only one tie. Against SOA, it maintained strong dominance with seven significant improvements, while showing more balanced competition with WO, though still maintaining an overall advantage. These patterns suggest that MWO's expert-guided strategy and adaptive search mechanisms contribute to robust performance across different optimization scenarios.

These results statistically validate that MWO demonstrates superior or competitive performance against state-of-the-art algorithms across different types of optimization landscapes, with particularly strong performance against SCSO and PEOA where it never showed inferior performance.
\section{Conclusion}

This paper has presented a novel Memetic Walrus Optimizer (MWO) to address the Adaptive Curriculum Sequencing challenge. This research advances the field in three key aspects. First, we developed an expert-guided optimization framework that synergizes the exploration capabilities of the walrus optimizer with expert knowledge for enhanced exploitation. Second, we designed a dynamic control mechanism that adaptively balances the global search and local refinement phases, leading to more stable convergence behavior. Third, we developed a comprehensive three-tier priority mechanism that successfully addresses multiple educational constraints while generating personalized learning sequences.

Comprehensive evaluations on the OULAD dataset validate the effectiveness of our MWO method. The results show that MWO not only achieves better optimization performance in terms of solution quality and stability but also generates more educationally meaningful learning sequences compared to existing methods. Additionally, benchmark function tests further validate the algorithm's robust optimization capability across different scenarios.

Future work could focus on extending the current framework to handle more dynamic educational scenarios, such as real-time student feedback and evolving learning objectives.Another promising direction is to explore deep learning techniques for enhancing expert knowledge representation and adaptation mechanisms.
\section*{Acknowledgements}

\bibliographystyle{elsarticle-num}
\bibliography{sample}

\end{document}